\documentclass[conference]{IEEEtran}

\usepackage{hyperref} 
\usepackage{booktabs} 

\usepackage[ampersand]{easylist}
\usepackage[font=small,skip=5pt]{caption}

\IEEEoverridecommandlockouts

\usepackage{cite}
\usepackage{amsmath,amssymb,amsfonts}
\usepackage{algorithmic}
\usepackage{graphicx}
\usepackage{textcomp}
\usepackage{xcolor}
\usepackage{hyperref}

\def\BibTeX{{\rm B\kern-.05em{\sc i\kern-.025em b}\kern-.08em
    T\kern-.1667em\lower.7ex\hbox{E}\kern-.125emX}}

\begin{document}

\title{What Lies Beneath: A Call for Distribution-based Visual Question \& Answer Datasets}

\author{\IEEEauthorblockN{Jill Naiman}
\IEEEauthorblockA{\textit{Information Science} \\
\textit{University of Illinois}\\
\textit{Urbana-Champaign}\\
Illinois, USA }
\and
\IEEEauthorblockN{Daniel J. Evans}
\IEEEauthorblockA{\textit{Information Science} \\
\textit{University of Illinois}\\
\textit{Urbana-Champaign}\\
Illinois, USA }
\and
\IEEEauthorblockN{JooYoung Seo}
\IEEEauthorblockA{\textit{Information Science} \\
\textit{University of Illinois}\\
\textit{Urbana-Champaign}\\
Illinois, USA }
%
}

\maketitle

\begin{abstract}
Visual Question Answering (VQA) has become an important benchmark for assessing how large multimodal models (LMMs) interpret images. However, most VQA datasets focus on real-world images or simple diagrammatic analysis, with few focused on interpreting complex scientific charts. Indeed, many VQA datasets that analyze charts do not contain the underlying data behind those charts or assume a 1-to-1 correspondence between chart marks and underlying data. In reality, charts are transformations (i.e. analysis, simplification, modification) of data. This distinction introduces a reasoning challenge in VQA that the current datasets do not capture.
In this paper, we argue for a dedicated VQA benchmark for scientific charts where there is no 1-to-1 correspondence between chart marks and underlying data. To do so, we survey existing VQA datasets and highlight limitations of the current field. We then generate synthetic histogram charts based on ground truth data, and ask both humans and a large reasoning model questions where precise answers depend on access to the underlying data. We release the open-source dataset, including figures, underlying data, distribution parameters used to generate the data, and bounding boxes for all figure marks and text for future research. 

\end{abstract}


Scientific charts are crucial for presenting quantitative information in academic articles. They allow readers to better comprehend patterns, compare values, and examine particularly notable aspects of a dataset \cite{tufte_visual_1983}. Recent advances in large multimodal models (LMMs) and large reasoning models (LRMs) have introduced the possibility of performing visual data interpretation through natural language interaction (e.g., GPT-5 \cite{openai2025introducinggpt5}). Through a combination of image recognition with language reasoning, it is possible to use these systems to answer questions about charts and plots. 

Visual Question Answering (VQA) has become an important benchmark for assessing the accuracy of these systems. 
To date, most VQA datasets devoted to scientific charts analyze an LMM's ability to answer varying factual-recall questions about the information on the page \cite{huang_pixels_2024}. 
However, many VQA datasets do not include access to the underlying data used to generate the figures or are missing bounding boxes for chart elements (e.g., axis labels), thus limiting future complex analysis with the datasets.  Additionally, when underlying data is included, the charts are limited to a 1-to-1 correspondence between chart mark and tabulated data.  As such, current VQA datasets lack coverage of \textit{distributions} where charts are not direct visual representations of the data.

To address this gap, we present a VQA benchmark for scientific charts focused on histogram charts where there is no 1-to-1 correspondence between chart marks and the underlying data. Our study surveys existing VQA datasets where we highlight the lack of distribution-based charts. We then generate a series of simple, synthetic histogram charts based on ground truth data and ask statistics-based questions of both two human annotators and an LMM \cite{openai2025introducinggpt5}. 
We release the preliminary open-source dataset, including figures, underlying data, distribution parameters used to generate the data, bounding boxes for all figure marks and text, and human and LMM answers to several questions for future research.

Our study is driven by the following research questions:

\begin{enumerate}
    \item To what extent do existing VQA datasets represent the reasoning demands of scientific charts, and where do they fall short?
    
    \item How accurately can LMMs answer questions about synthetic histogram charts when there is no direct link to the dataset? 
    
    \item How does the accuracy of an LMM compare to the accuracy of humans in answering statistical questions related to histograms?
    
\end{enumerate}

\section{Related Work in Scientific Chart VQA}


VQA is a research process by which humans can use AI to investigate information presented visually in figures through natural language queries \cite{agrawal_vqa_2016}. It codifies a series of tasks that are presented to an AI model and translate both multiple choice and free-form questions about visual objects into natural language answers. These questions vary in complexity while targeting specific aspects of an image \cite{vqainitial}. 

\begin{table*}
\begin{center}
\scriptsize
\caption{Summary of Chart VQA datasets with linked data. }
\label{tab:other_datasets}
\begin{tabular}{lllllllllll}
\toprule
Dataset Name & Year(s) & Line & Bar & Scatter & Pie & Histogram & Other & Bounding Boxes & Linked Data & Citation \\
\midrule
ChartAssistant/ChartSFT & 2024 & $\checkmark$ & $\checkmark$ & $\checkmark$ & $\checkmark$ & - & $\checkmark$ & - & partial & \cite{meng_chartassistant_2024} \\
Chart2Text/Chart-to-text & 2020/2022 & $\checkmark$ & $\checkmark$ & $\checkmark$ & $\checkmark$ & - & $\checkmark$ & partial & $\checkmark$ & \cite{kantharaj_chart--text_2022}\\
Chocolate & 2024 & $\checkmark$ & $\checkmark$ & - & - & - & - & - & $\checkmark$ & \cite{huang_lvlms_2024} \\
ChartQA/ChartQAPro & 2022/2025 & $\checkmark$ & $\checkmark$ & - & $\checkmark$ & - & - & $\checkmark$ & $\checkmark$ & \cite{masry_chartqa_2022} \\
ChartCheck & 2023 & $\checkmark$ & $\checkmark$ & $\checkmark$ & $\checkmark$ & - & $\checkmark$ & - & $\checkmark$ & \cite{akhtar_chartcheck_2024} \\
OpenCQA & 2022 & $\checkmark$ & $\checkmark$ & $\checkmark$ & $\checkmark$ & - & $\checkmark$ & $\checkmark$ & partial & \cite{kantharaj_opencqa_2022} \\
MapQA & 2022 & - & - & - & - & - & $\checkmark$ & $\checkmark$ & $\checkmark$ & \cite{chang_mapqa_2022} \\
VisText & 2023 & $\checkmark$ & $\checkmark$ & - & - & - & $\checkmark$ & partial & $\checkmark$ & \cite{tang_vistext_2023} \\
Chart-Info & 2019-2024 & $\checkmark$ & $\checkmark$ & - & - & - & $\checkmark$ & $\checkmark$ & partial & \cite{davila2022icpr} \\
DocVQA & 2021 & $\checkmark$ & $\checkmark$ & - & $\checkmark$ & - & - & - & $\checkmark$ & \cite{mathew_docvqa_2021} \\
ChartFC & 2023 & - & $\checkmark$ & - & - & - & - & $\checkmark$ & $\checkmark$ & \cite{akhtar_reading_2023} \\
RealCQA & 2023 & $\checkmark$ & $\checkmark$ & $\checkmark$ & - & - & $\checkmark$ & $\checkmark$ & $\checkmark$ & \cite{ahmed_realcqa_2023} \\
StructChart & 2023 & $\checkmark$ & $\checkmark$ & - & $\checkmark$ & $\checkmark$ & $\checkmark$ & - & partial & \cite{xia_structchart_2024} \\
LineEX & 2023 & $\checkmark$ & - & - & - & - & - & $\checkmark$ & $\checkmark$ & \cite{p_lineex_2023} \\
PlotQA & 2020 & $\checkmark$ & $\checkmark$ & $\checkmark$ & - & - & - & $\checkmark$ & $\checkmark$ & \cite{methani_plotqa_2020} \\
MatCha & 2023 & $\checkmark$ & $\checkmark$ & - & - & - & $\checkmark$ & - & $\checkmark$ & \cite{liu_matcha_2023} \\
TQA & 2017 & - & - & - & - & - & $\checkmark$ & $\checkmark$ & $\checkmark$ & \cite{kembhavi_are_2017} \\
\bottomrule
\end{tabular}
\end{center}
\end{table*}

More recently, VQA has been employed to benchmark visual comprehension of generative AI. The proliferation of large language models (LLMs) \cite{cai_leveraging_2023}, large vision models (LVMs) \cite{li_multimodal_2024, wang_review_2023} and LMMs  \cite{cai_vip-llava_2024} are prompting researchers to examine the role of AI in vision creation and comprehension. In doing so, VQA provides a framework for evaluating more complex images and images in context \cite{marino_ok-vqa_2019}.


More pointedly, researchers are using VQA to determine the ability of machine learning models to evaluate question-answering capabilities related to plots \cite{methani_plotqa_2020}, figures \cite{kahou_figureqa_2018}, and charts \cite{wu_dcqa_2023} within scientific articles using natural language reasoning using both synthetically generated \cite{lim_survey_2016} and real-world datasets (e.g., Statista and Pew, \cite{kantharaj_opencqa_2022,tang_vistext_2023}). The resulting question-answer datasets from these experiments are useful in evaluating new models or sets of documents \cite{sundar_cpapers_2024}. 



A summary of VQA datasets for charts is shown in \autoref{tab:other_datasets} indicating type of chart, whether or not bounding boxes for chart items (data marks, axis labels, etc.) are included.  For brevity, we limit \autoref{tab:other_datasets} to datasets which have some form of linked data.
Bounding boxes marked ``partial'' indicate some but not all elements are marked, and linked data marked as ``partial" indicates data is available for some but not all charts within the dataset.  We note \autoref{tab:other_datasets} refers to the \textit{publicly available} bounding boxes/linked data as several VQA datasets use underlying datasets which are not available publicly (e.g., ExcelChart400K, \cite{luo_chartocr_2021}). 


Our research found that histograms are an underrepresented chart type in the current VQA corpus. Although some VQA datasets include histograms as part of their analysis \cite{liu_mmc_2024,xia_structchart_2024}, few if any datasets focus specifically on them, and when present, links to original data are missing \cite{liu_mmc_2024} or shared data are already binned \cite{xia_structchart_2024}, even when charts are tied to real-world data \cite{kantharaj_opencqa_2022,tang_vistext_2023}. In the absence of a corpus of scientific documents, we propose a framework for generating synthetic data for VQA analysis.

\section{Figure \& Question Dataset Generation}

To test the VQA abilities of humans and LMMs, we generate synthetic scientific figures 
\cite{kahou_figureqa_2018,methani_plotqa_2020,wu_dcqa_2023}
using \textsf{Python}'s \textsf{matplotlib} package.  While training models with synthetic data can lead to decreased generalizability on real world tasks \cite{huang_pixels_2024}, for the quantification of LMM accuracy on simple VQA tasks with the distribution-based data presented here, synthetic data allows for a more controlled study.

The released code supports the generation of several types of plots (histogram, line, scatter, and contour plots), and several types of underlying model distributions (linear, random, Gaussian mixture models). The code also allows for variations in axis labels, titles, colors, font, scales (log/linear), errorbars, DPI, aspect ratio, and number of plots per figure.  

\begin{figure*}
\centering
\includegraphics[width=2\columnwidth]{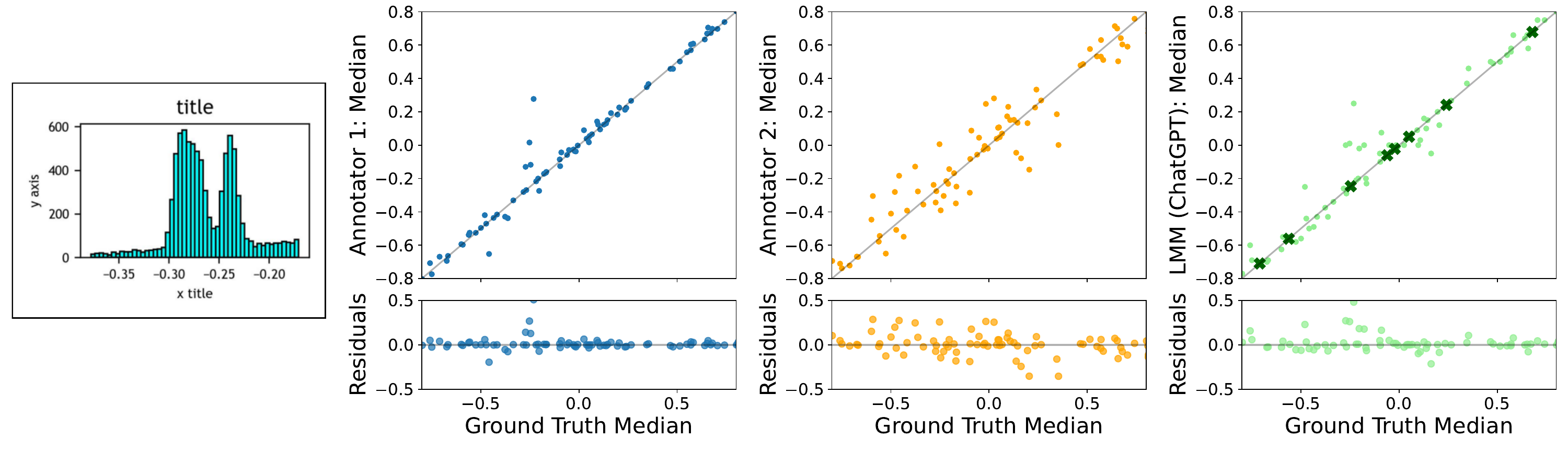} 
\caption{Example histogram (left) along with results of human annotators (second from left, second from right) vs LMM (right) estimations of median of underlying histogram data across 80 histograms.  Residuals from the ground truth answers are shown in bottom panels.  The green ``X"'s in the LMM plot denotes a incorrectly formatted response from the LMM at the specified ground truth median (LMM refuses to answer, or median is approximately an order of magnitude outside plotted range). 
}
\label{fig:ann_med}
\end{figure*}

However, to reduce complexity in this work, we fix all but the parameters of the Gaussian mixture models --  the number of gaussians is chosen between 1-5, and a range of data points for histograms to values in (-1,1), with additional random noise added between 5-10\%.  Along with randomly generated figures, our pipeline generates questions for specific panels in a one- to several-panel figure, however, we restrict this work to single-panel figures. 
The left panel of \autoref{fig:ann_med} shows an example of one of the generated histograms.

Our images are formatted as JPEG, and the figure data and questions are JSON.  Our LMM prompt questions are motivated by the work of \cite{seo_maidr_2024}. These are distributed into the ``levels" \cite{lundgard_accessible_2022}, with higher-level questions indicating requiring more reasoning and complexity to answer.  Questions are stored in full, and also broken up into sentences by ``persona" (role of the LMM), ``context" (context in which the question is being answered), ``question" (simplest form of question) and ``format" (output format of answer) \cite{googleprompting} such that future users of the data can substitute their own personas/context/formats within the original question.

\section{Human \& LMM Prompting}

Using the generated histogram-based VQA data we address our second and third research questions by asking an LMM (ChatGPT-5) and two human annotators the following two questions (1) ``What is the median value of the data in this figure panel?" and (2) ``How many gaussians were used to generate the data for the plot in the figure panel?".
The addition of a constraint in the ``format" part of the prompt of ``Please choose an integer number from 1 to 5." was added to mimic the human annotator's knowledge that the number of gaussians was limited to the range 1-5. Tests performed without this limitation  increased the error rate of the LMM.

\begin{figure}
\centering
\includegraphics[width=1.0\columnwidth]{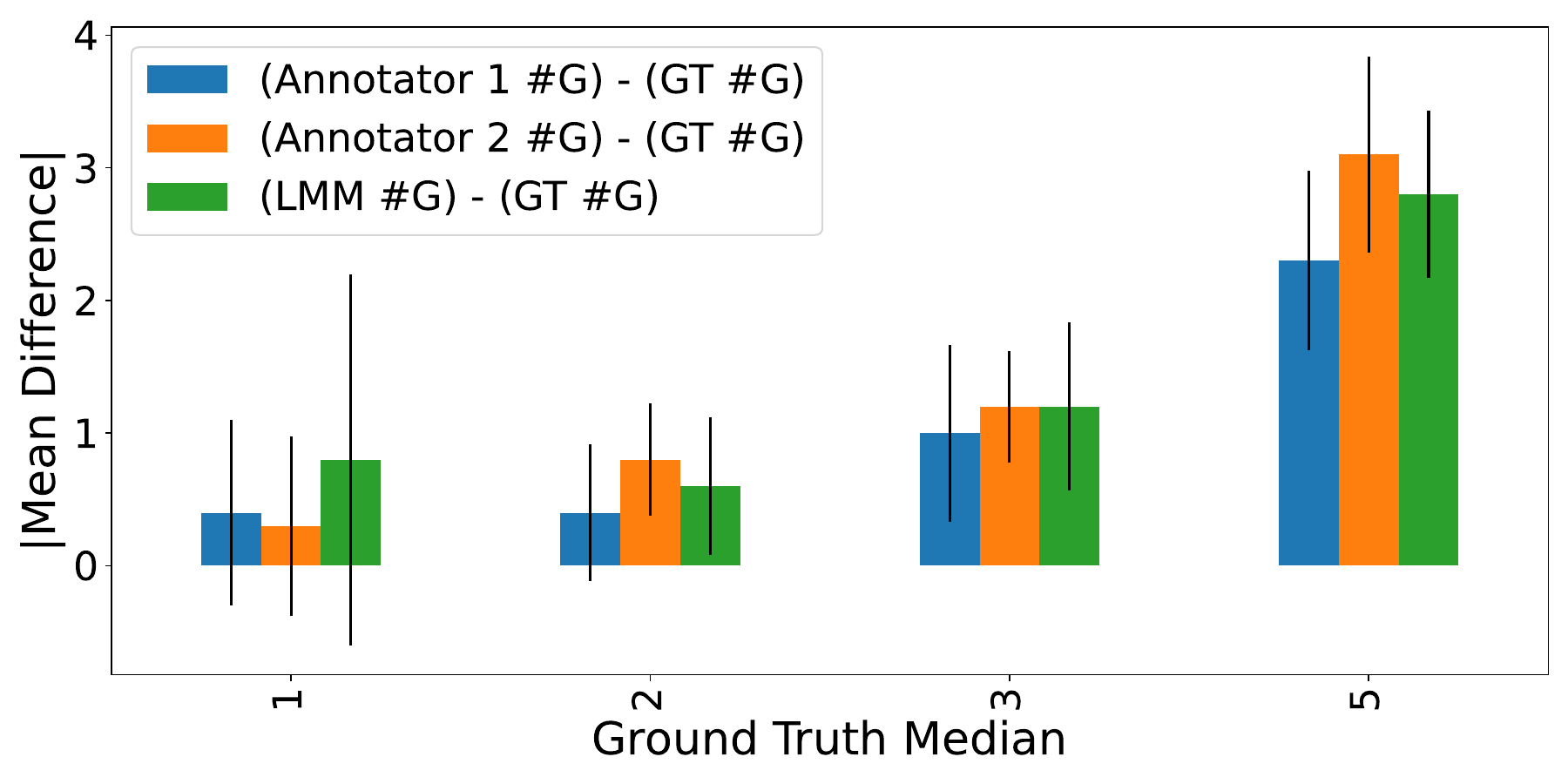} 
\caption{Human annotators and LMM estimate differences from ground truth (GT) for number of gaussians (\#G) in the underlying histogram distributions.}
\label{fig:ann_ngauss}
\end{figure}

To test the possible affects of the number of bars on accuracy rates, we fixed the number of gaussians to 2 and varied the bar size between values (10, 20, 45, 60) with 10 plots for each number of bars for a total of 40 histogram examples. 
To test the possible affects of the number of gaussians on accuracy, we fixed the number of bars to 50 and varied the number of gaussians between (1, 2, 3, 5) with 10 plots for each number of gaussians for a total of 40 additional histograms.

 Human annotations were made with the Zooniverse platform \cite{zooniverse} in which annotators were prompted with the histogram image and asked to input their numerical estimation for the median and number of gaussians. Annotators also "drew" the location of the median on the histogram as a vertical line.  
 The line location was translated onto the x-axis using the bounding boxes stored in our dataset. We found that annotation through drawing is more accurate for median estimations, thus in what follows when human measurements for median are discussed they refer to the ``drawn" annotation.

\section{Results and Discussion}

The results of the estimation of the median by two human annotators (right, middle panels) and LMM (gpt-5-nano, left panels) are shown in \autoref{fig:ann_med}.
All annotators show general agreement with the ground truth (solid lines), and residuals (lower panels) are generally clustered around the ground truth with no obvious trends present.  However, residual shape is more peaked and has longer tails across all annotators than a normal distribution, and all residuals fail a Shapiro-Wilk test \cite{shapiro_analysis_1965} for normalcy.

Thus, to determine if the estimated medians from the three groups differ from each other, we use the non-parametric Kruskal-Wallis test  \cite{kruskal1952use} which indicates there is \textit{no significant difference between the median} residuals across all three groups.
To test if there is significant difference between the variances of each residual distribution, we perform a Levene's test \cite{levene1960robust} and find there \textit{is significant difference between group variances}.

Doing a pair-wise Levene's test with the Bonferroni-Holm correction \cite{holm1979simple} we find that Annotator 2's residuals have a different variance than both Annotator 1 and the LMM.

This result aligns with the results presented in \autoref{fig:ann_med} where Annotator 2's variance appears larger than both Annotator 1 and the LMM.
This discrepancy could be due to the difference in backgrounds between Annotator 1 and 2 -- Annotator 1 has a background teaching statistics courses while Annotator 2 does not.
While similarities are apparent between Annotator 1 and LMM residuals, we reserve any judgment about the equivalent ``background" of the LMM in comparison to Annotator 1 and 2 to a more extensive study.

Next, we test the effects of number of histogram bars on estimates of the median by examining the subset of 40 histograms with a fixed number of gaussians of 2 and variable bin size of $n_{\rm bars} = [10, 20, 45, 60]$.  
Using a Kruskal-Wallis test, there are no difference in median or varience of residuals between annotators in any of the $n_{\rm bars}$ bins.

Finally, we use the subset of 40 histograms with a fixed bin size of 50 and test the impact of number of gaussians on annotator estimates. 
Using the Kruskal-Wallis test again, for each of our four ``number of gaussian" (\#G) bins, we find \textit{no significant} difference between the median estimations of either human or LMM annotator.
This aligns with \autoref{fig:ann_ngauss} which shows annotator estimates across the four choices of \#G -- in a given ground truth \#G bin the difference between mean estimated number of gaussians between human annotators (blue, orange) and LMM (green) is well within the errorbars for each bin (black lines).
However, \autoref{fig:ann_ngauss} does suggest a relationship between \#G and the ability of all annotators to estimate the number, with human and LMM errors increasing for an increasing \#G.  
To test this observation we use a ``linear mixed effects model" \cite{lindstrom1988newton} and find that when the effects of different annotators are included, there still \textit{is a significant} increasing error rate on estimation of \#G with increasing ground-truth \#G.
Intuitively this is not unexpected -- as more gaussians are added, blending effects would make separating their affects on distributions visually more difficult.

\section{Caveats, Conclusions, and Future Work}

In this work, we have created and tested a distribution-based VQA dataset and found for a small subset of questions, answers from a LMM (\textsf{GPT-5}) more closely resemble those of a human with a statistics background than a human without.  

This work is intended as a rallying cry for the creation and testing of ``distribution-based" VQA datasets, in which the associated chart marks do not contain a 1-to-1 correspondence to the data with which they were created. While the dataset presented in this paper is preliminary, we hope it draws light to an underrepresented type of dataset and showcases the abilities of LMMs to match, and sometimes surpass, human annotators in estimating answers to statistical questions.

To minimize complexity, we focused on histograms with a small set of parameters (e.g., data range) and only two VQA questions. Avenues for future work will include fine-tuned models, additional questions, various histogram data ranges and styles, along with all bounding boxes for text and marks.

Finally, our results rely on GPT-5-nano, the smallest of the GPT-5 models.  Our initial tests with larger models, such as GPT5-mini, found no improvement and increased hallucinations.  We include these results in our data and code at \url{https://github.com/ReadingTimeMachine/LLM_VQA_JCDL2025}


\bibliographystyle{IEEEtran}
\bibliography{./IEEEabrv,./references_short}


\end{document}